\documentclass{article}
 \usepackage[preprint]{log_2022} 

\usepackage{booktabs}            
\usepackage{multirow}            
\usepackage{amsfonts}            
\usepackage{graphicx}            
\usepackage{duckuments}          

\usepackage{algorithmic}
\usepackage{algorithm}

\usepackage{svg}
\usepackage{booktabs}
\usepackage{adjustbox}
\usepackage{wrapfig}
\usepackage{subcaption}

\usepackage{amsthm}

\usepackage[numbers,compress,sort]{natbib}

\title[Layer-wise training for self-supervised learning on graphs]{Layer-wise training for self-supervised learning on graphs}

\author[Pina et al.]{%
Oscar Pina\\
\institute{Universitat Politècnica de Catalunya - BarcelonaTech (UPC)}\\
\email{oscar.pina@upc.edu}\And
Verónica Vilaplana\\
\institute{Universitat Politècnica de Catalunya - BarcelonaTech (UPC)}\\
\email{veronica.vilaplana@upc.edu}
}

\begin{document}

\maketitle

\begin{abstract}


End-to-end training of graph neural networks (GNN) on large graphs presents several memory and computational challenges, and limits the application to shallow architectures as depth exponentially increases the memory and space complexities. In this manuscript, we propose Layer-wise Regularized Graph Infomax, an algorithm to train GNNs layer by layer in a self-supervised manner. We decouple the feature propagation and feature transformation carried out by GNNs to learn node representations in order to derive a loss function based on the prediction of future inputs. We evaluate the algorithm in inductive large graphs and show similar performance to other end to end methods and a substantially increased efficiency, which enables the training of more sophisticated models in one single device. We also show that our algorithm avoids the oversmoothing of the representations, another common challenge of deep GNNs.

\end{abstract}
\section{Introduction}
\label{sec:intro}

Graph neural networks (GNN) learn high order node representations by iteratively propagating and transforming node features. At every convolutional layer, a node aggregates and projects the information from its one-hop neighborhood. This procedure defines a tree-like computational graph for every node, which grows exponentially with the depth of the GNN. Therefore, end-to-end training of graph neural networks on large graphs presents several memory and computational challenges, limiting the application to shallow GNNs.

Traditionally, sampling strategies \cite{NIPS2017_5dd9db5e} are carried out in order to reduce the exponentially growing size of the nodes' receptive field. However, it comes at expense of a loss of information if the amount of samples is not sufficiently large. The most widely used methodology is neighborhood sampling, which leverges the tree-like computational graph and samples fixed-size neighborhoods for every node for each convolutional layer of the original architecture. 

Deep GNNs present other challenges characteristic of graph learning. For instance, as the number of layers increases, the exponentially growing neighborhood can lead to an oversmoothing of the representations, in which node embeddings are very similar to each other so that it is unfeasible to distinguish between them.

To overcome the aforementioned challenges, we propose an algorithm to train GNNs layer by layer in a self-supervised manner. Layer-wise training reduces the tree-like computational graphs to a one-hop graph, so that the memory and computational constraints are avoided. We decouple the feature propagation and feature transformation carried out by GNNs at each convolutional layer to learn node representations and derive a loss function based on predictive coding (PC), a biologically plausible framework to train neural networks locally rather than from a global loss function. According to PC, we train every layer of the GNN by maximizing the mutual information between output node representations and their propagation through the graph, which will be the input of the next layer. This objective function leads to a layer-wise extension of Regularized Graph Infomax \cite{pina2023rgi}, so that we call our algorithm \textit{LRGI: Layer-wise Regularized Graph Infomax}.

We show that our algorithm is able to efficiently fit GNNs on large graphs by training layer-wise while preserving the performance of its end-to-end counterpart. We quantify the memory and computational benefits of layer-wise training, and also demonstrate that our algorithm is able to avoid the oversmoothing of the representation despite the depth of the encoder leading to meaningful, useful representations. Our contributions are:

\begin{itemize}
    \item We propose the algorithm \textit{LRGI: Layer-wise regularized graph infomax}, a self-supervised learning algorithm for graphs. LRGI trains the distinct layers of a GNN encoder as independent modules employing the RGI \cite{pina2023rgi} algorithm.
    
    \item We argument and empirically show that LRGI reduces the training time and memory complexities which enables to train faster and increase the model's capacity in one single device.

     \item We show that layer-wise training also overcome other challenges of deep GNNs, such as vanishing gradients and oversmoothing.
\end{itemize}

This article is organised as follows, we first provide some background on PC for other deep learning domains, sampling strategies for GNNs and self-supervised learning on graphs in Section \ref{sec:related}. Then, we motivate and detail our algorithm LRGI and provide insights into its benefits in Section \ref{sec:method}. In Section \ref{sec:exp} we evaluate the performance and training efficiency of LRGI. Finally, we address the limitations and conclude the work in Section \ref{sec:limitations} and Section \ref{sec:conclusions}, respectively.

\section{Background and related work}
\label{sec:related}

\subsection{Predictive coding for neural network training} 
\label{sec:related_cpc}

Predictive coding is a theoretical framework used in neuroscience to explain how the brain processes sensory information, learns and makes predictions about the external world. It proposes that the brain generates predictions about future incoming sensory data and learns by minimizing the error signal between the prediction and the actual sensory inputs it receives. Contrastive Predictive Coding (CPC) \cite{oord2018representation} and Deep InfoMax (DIM) \cite{hjelm2018learning} are self-supervised learning frameworks that capture invariant features useful for downstream tasks. CPC focuses on maximizing the mutual information between temporally close (audio) or spatially adjacent (image) patches, that is, between neighbors, whereas DIM targets the information shared between local patches and the global structure of an image. Greedy InfoMax (GIM) \cite{lowe2019putting} bases on neuroscience evidence that the predictive coding learning in the brain is carried out locally rather than from a global objective to extend CPC by training a neural network layer by layer and learning features that are predictive of its future inputs \cite{lowe2019putting}.

\subsection{Sampling strategies for graph neural networks}
\label{sec:related_sampling}
Graph neural networks (GNN) iteratively aggregate node features of local neighborhoods to update the central node's representation. This procedure implicitly defines a tree-like computational graph that captures the information propagation unique to each node. After $L$ convolutional layers, every node will have received information from its $L$-hop neighborhood. Therefore, the number of neighbors needed to obtain the final representation for a given node grows exponentially with respect to the depth of the encoder. In large graph scenarios where mini-batching is required, a sampling strategy is defined to reduce the computational complexity and memory footprint. Neighbor sampling \cite{NIPS2017_5dd9db5e} consists of sampling fixed-size neighborhoods for every layer of the architecture and node of the mini-batch. Taking the full size neighborhoods without sampling, the worst case space complexity for one single batched node is $\mathcal{O}(N)$, where $N$ is the number of nodes of the graph. Instead, with neighbor sampling, this complexity is reduced to $\mathcal{O}  ( \prod_{l=1}^{L} S_{(l)}  )$, where $S_{(l)}$ is the number of nodes sampled for layer $(l)$, so that the final complexity can be controlled by tuning the hyperparameters $S_{(l)}$.

\subsection{Self-supervised learning on graphs}
\label{sec:related_gssl}
Early work in self-supervised graph representation learning is based on contrastive learning \cite{velickovic2018deep, icml2020_1971, peng2020graph} extending works from other domains such as computer vision \cite{hjelm2018learning, chen2020simple}. Deep Graph Infomax (DGI) \cite{velickovic2018deep} trains a graph neural network encoder by maximizing the mutual information between node-level (local) and graph-level (global) views. The algorithm targets node-level tasks, in which only one graph is available. The negative pairs are constructed by applying a random permutation on the node features and obtaing alternative graph-level representations. Other methods train the encoder to be invariant to random graph transformations \cite{Zhu:2020vf, thakoor2021bootstrapped,bielak2021graph,zhang2021canonical}. This is accomplished by obtaining two views of the same graph by applying these random transformations and then forcing the encoder to output similar representations for the two views. However, contrastive learning usually requires a large number of negative samples to perform competitively. Instead, BGRL \cite{thakoor2021bootstrapped} adopts a negative sampling-free scheme by constructing a teacher-student asymmetric architecture and training tricks. G-BT \cite{bielak2021graph} and CCA-SSG \cite{zhang2021canonical} employ a more intuitive method to prevent the encoder outputting constant representations: regularization on the covariance matrix of the representation space. The invariance approach includes strong assumptions, as it is assumed that the downstream task is also invariant to these transformations. Nonetheless, it has been argued that transformations such as edge sampling modify the graph semantics so the assumptions may be incorrect in particular domains \cite{icml2020_1971, NEURIPS2021_ff1e68e7}. Alternatively, methods leveraging graph diffusion \cite{icml2020_1971}, or the local structure of the graph data \cite{peng2020graph}, between others, have been proposed.

\subsection{Regularized graph infomax (RGI)}
\label{sec:related_rgi}
Regularized graph infomax (RGI) \cite{pina2023rgi} trains a graph neural network (GNN) encoder in a self-supervised manner by propagating the embeddings output by the encoder as self-supervision signals via mutual information maximization. Formally, for a graph of $N$ nodes, being $\mathbf{X} \in \mathbb{R}^{N \times d}$ the $d$-dimensional node features, $\mathbf{A} \in \mathbb{R}^{N \times N}$ the adjacency matrix, $\mathbf{U}=f_{L,\Theta}(\mathbf{X},\mathbf{A})$ the node embeddings output by the encoder and $\mathbf{V}=g_K(\mathbf{U},\mathbf{A})$ their propagation through the graph, RGI optimizes the loss function:

\begin{equation}
    \label{eq:loss_u}
    \mathcal{L}_u = \frac{\lambda_1}{N} \sum_{i=1}^{N} \parallel \mathbf{u}_i - h_{\phi}( \mathbf{v}_i ) \parallel_2^2 + \frac{\lambda_2}{D} \sum_{n=1}^{D} \left ( 1 - \mathcal{C}_{nn} \right )^2 +\frac{\lambda_3}{D} \sum_{n=1}^{D} \sum_{m \neq n} \mathcal{C}_{nm}^2
\end{equation}

where $h_{\phi}$ is a fully-connected neural network parametrized by $\phi$, which is jointly trained with the encoder to predict the node embedding $\mathbf{u}_i$ from the propagated embedding $\mathbf{v}_i$, $\mathcal{C}$ is the sample covariance matrix of $\mathbf{U}$, and $\lambda_1, \lambda_2, \lambda_3 \in \mathbb{R}$ are weight loss hyperparameters. The first term is the reconstruction error between the two representations, which promotes context-aware embeddings, whereas the second and third terms are the variance-covariance regularization proposed in \cite{bardes2022vicreg} to avoid the collapse of the representations. In practice, this loss is symmetrized by predicting $\mathbf{v}_i$ form $\mathbf{u}_i$ with another auxiliary network $h_{\psi}$ and regularizing the covariance matrix of the propagated embeddings. A complete explanation of the algorithm is provided in Appendix \ref{sec:app_rgi}.

\section{Method}
\label{sec:method}

\begin{algorithm}[tb]
   \caption{Layer-wise RGI}
   \label{alg:lrgi}
\begin{algorithmic}   \STATE {\bfseries Input:} node features $\mathbf{X}$; adjacency matrix $\mathbf{A}$; backbone $f_{L, \Theta}$; global view function $g_{K}$; reconstruction networks $\{h_{(l),\phi}\}_{l=1}^{L}$ and $\{h_{(l),\psi}\}_{l=1}^{L}$; propagation steps $K$.
   \STATE // use node features as first input
   \STATE $\mathbf{U}_{(0)} \leftarrow \mathbf{X}$
   \STATE // fit layers one after the other
   \FOR{$l=1...L$}
    \STATE // apply RGI to fit layer (l)
    \STATE $\mathbf{U}_{(l)} \leftarrow RGI(\mathbf{U}_{(l-1)}, \mathbf{A}, f_{(l),\Theta}, g_{K}, h_{(l),\phi}, h_{(l),\psi})$
   \ENDFOR
   \STATE {\bfseries return} $\mathbf{U}_{(L)}$
\end{algorithmic}
\end{algorithm}

In this section, we first theoretically motivate the layer-wise training of graph neural networks and the usage of RGI algorithm in Section \ref{sec:method_motivation}, then, we detail our algorithm \textit{LRGI - Layer-wise regularized graph infomax} in Section \ref{sec:method_lrgi} and finally provide extra practical advantages of the method in Section \ref{sec:method_advantages}. 

\subsection{Layer-wise training on graphs}
\label{sec:method_motivation}


We propose layer-wise training as a solution to overcome computational and memory challenges of end-to-end training of graph neural networks and rely on self-supervised learning to do so. We base our solution on the observation that a $L$ layer GNN encoder, denoted as $f_{\Theta}$, can be seen as a sequence of $L$ local aggregation and projection operations which could be trained locally, layer by layer. As predictive coding provides a framework for local training of neural networks based on neuroscience, we extend this approach to graphs and show that RGI is a good proxy. 


\paragraph{$f_{\Theta}$ as a sequence of $L$ local aggregations} GNNs iteratively perform feature aggregation and transformation in order to learn node representations. Concretely, to compute node $i$'s representation at layer $(l)$, the network employs the aggregation of its one-hop neighborhood after the previous convolutional layer, which can be expressed as:

\begin{equation}
    \label{eq:f_l}
    \mathbf{u}_{(l),i} \leftarrow f_{(l),\Theta}(\mathbf{u}_{(l-1),i}, AGG \{ \mathbf{u}_{(l-1),j} | j \in \mathcal{N}_i \})
\end{equation}

where $AGG$ is a permutation invariant aggregation function. After $L$ layers of graph convolutions, the node representations encode their $L$-hop neighborhood. A key insight driving our work is the realization that if each one of the inner layers $f_{(l),\Theta}$ exhibits expressiveness, the overall network will show this property as well.

\paragraph{Predicting the next hop} Predictive coding states that the brain learns locally by minimizing the prediction error of future inputs rather than from a global loss. Although we do not define a notion of \textit{future} inputs for graph structured data, we target what will be the input of the next layer: the next hop, or aggregation of one-hop neighbors. Formally, given the node embeddings obtained according to Equation \ref{eq:f_l}, the output of the current layer is $\mathbf{u}_{(l),i}$, whereas the input to the next layer $f_{(l+1),\Theta}$ to update node $i$'s representation will be $AGG \{ \mathbf{u}_{j,(l)} | j \in \mathcal{N}_i \}$. Therefore, under the predictive coding framework, the training of the layer $f_{(l),\Theta}$ can be addressed with the maximization of: 

\begin{equation}
    \label{eq:objective}
    I(\mathbf{u}_{(l),i}; AGG \{ \mathbf{u}_{(l),j} | j \in \mathcal{N}_i \})
\end{equation}

Intuitively, $AGG \{ \mathbf{u}_{(l),j} | j \in \mathcal{N}_i \}$ contains information of the next hop of node $i$, that is, about unseen nodes for $\mathbf{u}_{(l),i}$, so that this objective function would force the node representation to be able to predict future hops. RGI \cite{pina2023rgi} provides a framework to efficiently maximize the mutual information between node embeddings ($\mathbf{U}$) and their propagation through the graph ($\mathbf{V}$) for $K$ steps, which encode the local and global context of the nodes in the graph, respectively. Fixing the value $K=1$, we obtain the proposed objective function. This idea is developed and the RGI algorithm is extended to our layer-wise setting in Section \ref{sec:method_lrgi}.


\subsection{Layer-wise RGI}
\label{sec:method_lrgi}

In order to train every $f_{(l),\Theta}$ based on the expression of Equation \ref{eq:objective}, we introduce \textit{LRGI : Layer-wise Regularised Graph Infomax}, a layer-wise, self-supervised learning algorithm that locally trains each layer $f_{(l),\Theta}, l \in \{1,...,L\}$ of a GNN encoder. This is achieved by employing a particular case of RGI \cite{pina2023rgi} algorithm (see Section \ref{sec:related_rgi}) and applying it independently to every layer. To match the objective function of Equation \ref{eq:objective}, we must set $K=1$ to only incorporate one-hop neighbors.

In the layer-wise scenario, following the original algorithm's notation, we define local and global embeddings for every layer $(l)$, denoted as  $\mathbf{U}_{(l)} = f_{(l),\Theta}(\mathbf{U}_{(l-1)}, \mathbf{A})$ and $\mathbf{V}_{(l)} = g_{K}(\mathbf{U}_{(l)}, \mathbf{A})$, respectively. We implement $g_K$ by setting $K=1$ and employing the degree normalized adjacency matrix to propagate the embeddings rather than the symmetrically normalized matrix proposed in the original paper, so that  $\mathbf{v}_{(l),i} = AGG \{ \mathbf{u}_{(l),j} | j \in \mathcal{N}_i \} = \sum_{j \in \mathcal{N}_i} \mathbf{u}_{(l),j} / d_i $. In addition, although the same architecture is employed, the weights of the auxiliary reconstruction networks $h_{(l),\phi}$ and $h_{(l),\psi}$ are different for every layer, and they are removed after training. The algorithm is described in Algorithm \ref{alg:lrgi}.

\begin{figure}
	\centering
	\includegraphics[width=\textwidth]{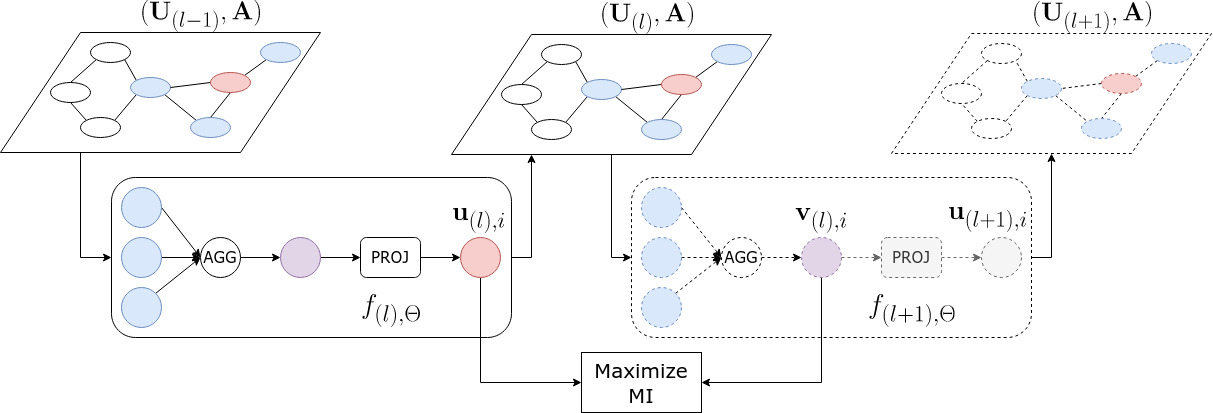}
	\caption{Visual illustration of LRGI.}
	\label{fig:lrgi}
    \small The layer $f_{(l),\Theta}$ is trained by maximizing the mutual information between the output node representations $\mathbf{u}_{(l),i}$ and the input to the next layer, that is, the aggregation of the one-hop neighbors: $\mathbf{v}_{(l),i} = AGG \{ \mathbf{u}_{(l),j} | j \in \mathcal{N}_i \}$
\end{figure}

\subsection{Practical advantages}
\label{sec:method_advantages}

\paragraph{Computational benefits} As it has been stated, layer-wise training is especially advantageous in scenarios involving large graphs where batching and sampling are required. Following the neighbor sampling strategy \cite{NIPS2017_5dd9db5e}, each batch comprises a subset of target nodes and its sampled $L$-hop neighborhood for which intermediate representations are computed, but the loss function is only applied to the target nodes. Layer-wise training, instead, decouples the amount of sampled nodes per batch from the depth of the encoder, that is, the per-batch space and memory complexities are reduced from $\mathcal{O}  ( \prod_{l=1}^{L} S_{(l)}  )$ to $\mathcal{O}  ( S_{L} )$, assuming that a fixed-size neighborhood ($S_{L}$) is sampled for every layer. However, training $f_{(l),\Theta}$ employing LRGI requires a two-hop neighborhood, as not only the target node representations but also their neighbors are needed to propagate them and create the self-supervision signals, so the actual complexity for LRGI is $\mathcal{O}  (S_{L} S_{K})$, where $S_{K}$ is the amount of sampled neighbors to compute the propagated embeddings, $S_{K}<S_{L}$.

\paragraph{Other training benefits} Vanishing gradients is a common problem faced when training deep neural networks in which gradients at initial layers vanish when backpropagated from deeper layers. Nonetheless, LRGI trains every layer locally, so that gradients are computed independently for every layer and no backpropagation through deep layers is required, overcoming then the vanishing gradients.

\paragraph{Representation benefits} The efficiency of layer-wise training provides more flexibility when designing GNN architectures as computational and memory constraints are overcome. For instance, the depth and dimensionality of the architectures can be increased, giving more representation power to the model. Additionally, oversmoothing is another well known problem of GNNs, in which if the network is too deep, node representations tend to be similar, or oversmoothed, so that it becomes unfeasible to carry on discrimination tasks between nodes. However, with LRGI, variance-covariance regularization is applied to the output of every layer, forcing the layers to output spread out, uncorrelated node representations so that oversmoothing is avoided. This statement is empirically shown in Section \ref{sec:exp_oversmoothing}.

\section{Experiments}
\label{sec:exp}

In this section we evaluate our algorithm in multiple graph benchmarks. In Section \ref{sec:exp_results} we compare LRGI with other state-of-the-art methods in terms of performance on downstream tasks under the linear evaluation protocol. Then, we also evaluate the space and time complexity of these algorithms in Section \ref{sec:exp_complexity} and based on the efficiency of LRGI we show that either the batch complexity or the model's capacity can be increased in Section \ref{sec:exp_batch} and Section \ref{sec:exp_capacity}, respectively. Finally, we show that LRGI also avoids the oversmoothing of the representations in Section \ref{sec:exp_oversmoothing}.

\paragraph{Inductive learning} Current works on self-supervised learning on graphs mainly focus on transductive settings, requiring all nodes of the graph to be present during training. This limits the landscape of self-supervised learning on graphs, as the goal of self-supervised learning is to fit a function that output useful representations to carry out downstream tasks even for unseen data. In order to better assess how self-supervised learning algorithms for graphs generalize, we focus on inductive learning settings. Although some of the graph bencharks we are using are proposed for transductive scenarios, we perform our experiments under an inductive setting, which is closer to real world scenarios, are generally more challenging and the performance gap with respecto to supervised learning is wider.


\paragraph{Datasets} We evaluate our algorithm on four popular graph bencharks under an inductive setting: \textit{PPI}, \textit{Reddit} and \textit{ogbn-products}. Table \ref{tab:datasets} shows the size and statistics of the datasets. 

\paragraph{Linear evaluation} Based on previous works of self-supervised learning on graphs \cite{velickovic2018deep}, we rely on the linear prove to assess the quality of the learned representations. As we are dealing with inductive settings, we train the encoder in a self-supervised manner with exclusively the training nodes' induced subgraph, in except of the PPI dataset, for which we also include the validation graphs. Then, we obtain the training node representations using the corresponding subgraph and finally the validation and test node embeddings using the entire graph. A linear classifier is fit and evaluate on the downstream task without backpropagating the gradients through the encoder.

\begin{table}[ht]
    \centering
    \caption{Datasets statistics}
    \label{tab:datasets}
    \small
        \begin{tabular}{l|ccccc}
        \toprule
        \textbf{Dataset} & \textbf{\#Graphs} & \textbf{\#Nodes} & \textbf{\#Node features} & \textbf{\#Edges} & \textbf{\#Classes} \\
        \midrule
    
        \textbf{PPI} & 24 & 2,245 & 50 & 61,318 & 121 (ML) \\
        \textbf{Reddit} & 1 & 232,965 & 602 & 114,615,892 & 41 \\
        \textbf{ogbn-products} & 1 & 2,449,029 & 100 & 61,859,140 & 47 \\
        \bottomrule
        \end{tabular}
\vskip 0.1in
\end{table}

\paragraph{Architecture} As proposed in previous works \cite{pina2023rgi, hou2022graphmae, bielak2021graph, thakoor2021bootstrapped}, the architecture we have employed for inductive settings is a composition of Graph Attention Network (GAT) \cite{velickovic2018graph} with linear skip connections and ELU activation. The default depth ($L$) and width ($D$) of the architecture, as well as the propagation steps ($K$), the batch size ($B$), sampling factor for the first $L$ hops ($S_L$) and sampling factor for the next $K$ hops ($S_K$) employed for the experiments of Section \ref{sec:exp_results} can be found in Table \ref{tab:hyperparams}. In order to simplify the architecture, we slightly modify related works' implementations and design every layer equally, being their output a $D$-dimensional feature vector divided into 4 heads.

\begin{table}[ht]
    \centering
    \caption{Default architecture hyperparameters of the reported experiments of Section \ref{sec:exp_results} }
    \label{tab:hyperparams}
    \small
        \begin{tabular}{l|cccccc}
        \toprule
        \textbf{Dataset} & $L$ & $D$ & $K$ & $B$ & $S_L$ & $S_K$ \\
        \midrule
        \textbf{PPI} & 3 & 1024 & 1 & 1 (graph) & - & - \\
        \textbf{Reddit} & 2 & 512 & 1 & 128 & 10 & 5 \\
        \textbf{ogbn-products} & 3 & 128 & 1 & 512 & 10 & 5 \\
        \bottomrule
\end{tabular}
\vskip 0.1in
\end{table}

\subsection{Numerical results}
\label{sec:exp_results}

Table \ref{tab:results} shows the performance comparison in terms of accuracy (micro-average F-Score for PPI dataset) of distinct state of the art methods under the linear evaluation protocol. LRGI achieves similar performance to state-of-the-art, and even outperforms other methods on the large scale \textit{ogbn-products}, despite it requires much less computation and memory resources to be trained, as shown in Section \ref{sec:exp_complexity}. Note that we do not add relevant works to the comparison table such as GRACE \cite{Zhu:2020vf}, BGRL \cite{thakoor2021bootstrapped} and CCA-SSG \cite{zhang2021canonical} since they are based on the same invariance via data augmentation principle than G-BT \cite{bielak2021graph}, but their objective function is not originally defined nor they provide the code for batched, inductive scenarios. GraphMAE \cite{hou2022graphmae} on \textit{ogbn-products} is not reported as the code is not prepared for batched scenarios and the full training subgraph leads to out-of-memory (OOM) error.

\begin{table}[ht]
\centering
\caption{Classification accuracy (Reddit and ogbn-products) and micro-average F1 score (PPI) averaged for 5 weight initializations.}
\label{tab:results}
\small
\begin{tabular}{l|ccc}
    \toprule
    \textbf{Method} & \textbf{PPI} & \textbf{Reddit} & \textbf{ogbn-products} \\
    \midrule  
    Raw ft. & 42.20 & 58.50 & 50.90 \\
    \midrule
    Rdm-Init & 62.60 $\pm$ 0.20 & 93.30 $\pm$ 0.00 & 39.91 $\pm$ 4.01 \\
    DGI \cite{velickovic2018deep} & 63.80 $\pm$ 0.20 & 94.00 $\pm$ 0.10 & 58.25 $\pm$ 3.70 \\
    G-BT \cite{bielak2021graph} & 74.13 $\pm$ 0.13 & 94.93 $\pm$ 0.09 & 69.47 $\pm$ 0.13 \\
    GraphMAE \cite{hou2022graphmae} & 74.50 $\pm$ 0.29 & \textbf{96.08 $\pm$ 0.08} & - \\
    RGI \cite{pina2023rgi} & \textbf{82.40 $\pm$ 0.09} & 95.82 $\pm$ 0.06 & 69.32 $\pm$ 0.57 \\
    \midrule
    LRGI (\textit{ours}) & 79.74 $\pm$ 0.25 & 95.44 $\pm$ 0.07 & \textbf{70.77 $\pm$ 0.53} \\
    \bottomrule
\end{tabular}
\vskip 0.05in
\end{table}

\subsection{Time and memory complexities}
\label{sec:exp_complexity}

\begingroup
\renewcommand{\arraystretch}{1.5} 
\begin{table}[ht]
\centering
\caption{Space and time complexities on \textit{ogbn-products}.}
\label{tab:complexities}
\small
\begin{tabular}{l|c|cc|c}
    \toprule
    \textbf{Method} & \textbf{Theroretical} & \textbf{\#Nodes} & \textbf{\#Edges } & \textbf{Time} \\
    \midrule
    Original & $\mathcal{O} ( \prod_{l=1}^{L} S_{(l)} )$ & 129,200 & 381,056 & - \\
    \midrule
    DGI \cite{velickovic2018deep} & $\mathcal{O} ( \prod_{l=1}^{L} S_{(l)} )$ & 129,200 & 381,056 & 01h15m \\
    G-BT \cite{bielak2021graph} & $\mathcal{O} (2 \times \prod_{l=1}^{L} S_{(l)} )$ & 258,400 & 762,112 & 01h30m \\
    RGI \cite{pina2023rgi} & $\mathcal{O} (S_K \times \prod_{l=1}^{L} S_{(l)} )$ & 169,862 & 834,140 & 01h24m \\
    \midrule
    LRGI (\textit{ours}) & $\mathcal{O} (S_K \times S_L )$ & \textbf{23,769} & \textbf{28,727} & \textbf{01h00m} \\
    \bottomrule
\end{tabular}
\vskip 0.05in
\end{table}
\endgroup

Table \ref{tab:complexities} shows the space complexity of a batch of one node of \textit{ogbn-products} for a network of $L=3$ layers and sampling $S_L=10$ nodes per layer. \textit{Theoretical} is the theoretical space complexity for a batch of one node, \textit{\#Nodes} and \textit{\#Edges} are the average number of nodes and edges sampled in a batch with batch size $B=512$ used for training as reported in Table \ref{tab:hyperparams}, and \textit{Time} is the training time for 100 epochs. For RGI and LRGI, we assume $K=1$ and $S_K=5$. We ommit GraphMAE \cite{hou2022graphmae} from the table as it is not originally defined for batched scenarios and the extension may be non-trivial. Augmentation-based methods such as G-BT \cite{bielak2021graph} do not increase the depth of the batch but they double its complexity, as they obtain two different graph views via random transformations. DGI \cite{velickovic2018deep}, instead, do not increase the complexity of the original raw batch. With layer-wise training, we only need to sample 1-hop neighborhoods, substantially reducing the batch complexity and memory footprint. Therefore, as the number of intermediate representations that are computed is reduced, the layer-wise training time is the lowest among the methods.

\subsection{Increasing the batch complexity}
\label{sec:exp_batch}

Neighbor sampling and mini-batching are two different sources of noise that may increase the variance and information loss during training. Moreover, small batch sizes also increases the training time. With layer-wise training, it is possible to get rid of neighbor sampling as the space complexity is still feasible, or also to increase the batch size to accelerate training. Table \ref{tab:batch} shows the effect of the batch size and the sampling factor on the training time when training layer-wise. $| \mathcal{N}_i |$ denotes full neighborhood sampling. As expected, increasing $B$ substantially reduces the training time, enabling training on large graphs in about half an hour. Layer-wise training makes it feasible to train with full-sized neighborhoods, although it comes at expense of an increased time complexity.

\begin{table}[ht]
\centering
\caption{Training time with larger batch size and/or no neighbor sampling.}
\label{tab:batch}
\small
\begin{tabular}{ccc|cc|c}
    \toprule
    $B$ & $S_L$ & $S_K$ & \textbf{\#Nodes} & \textbf{\#Edges } & \textbf{Time} \\
    \midrule
    512 & 10 & 5 & 23,769 & 28,727 & 01h00m \\
    512 & $| \mathcal{N}_i |$ & 5 & 73,377 & 152,894 & 02h15m \\
    2048 & 10 & 5 & 71,047 & 107,037 & 00h38m \\
    2048 & $| \mathcal{N}_i |$ & 5 & 141,932 & 483,078 & 00h53m \\
    \bottomrule
\end{tabular}
\vskip 0.05in
\end{table}

\subsection{Increasing models' capacity}
\label{sec:exp_capacity}
As layer-wise training decouples the batch space complexity from the depth of the encoder, it enables to increase its capacity but still being able to fit it in one single device. Additionally, self-supervised learning algorithms generally benefit from high dimensional embeddings \cite{bardes2022vicreg, grill2020bootstrap}, which also demands more memory.  
Figure \ref{fig:depth_width_ppi} shows the effect of the encoder's depth $(L)$ and width $(D)$ on the PPI dataset as micro-average F-score performance with the linear evaluation protocol. As it is expected, the dimensionality has a significant impact on the downstream performance for both RGI and LRGI. In Figure \ref{fig:depth_width_ppi} it can be observed that the widest model outperforms the supervised baseline whereas the models with lower dimensionalities are unable to capture all the necessary information to fit the downstream task. On the other hand, RGI converges faster than LRGI to its best performance when the number of layers is optimal, that is, between 3 and 5 layers. Note, however, that RGI with a dimensioanlity of $D=4096$ cannot be trained with a model deeper than $L=5$ due to out-of-memory error. Moreover, as the number of layers increases, RGI's performance drops for all dimensionalities, but LRGI's performance does not. This is due to the oversmoothing of the representations, as shown in Section \ref{sec:exp_oversmoothing}.

\subsection{Avoiding oversmoothing}
\label{sec:exp_oversmoothing}

\begin{figure}
  \centering
  \begin{subfigure}{0.48\textwidth}
    \includegraphics[width=\linewidth]{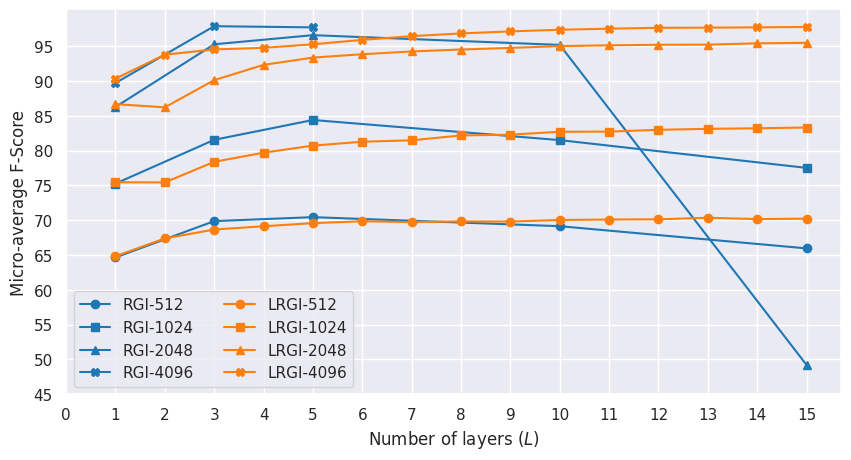}
    \caption{Effect of $D$ and $L$ on RGI and LRGI.}
    \label{fig:depth_width_ppi}
  \end{subfigure}
  \hfill
  \begin{subfigure}{0.48\textwidth}
    \includegraphics[width=\linewidth]{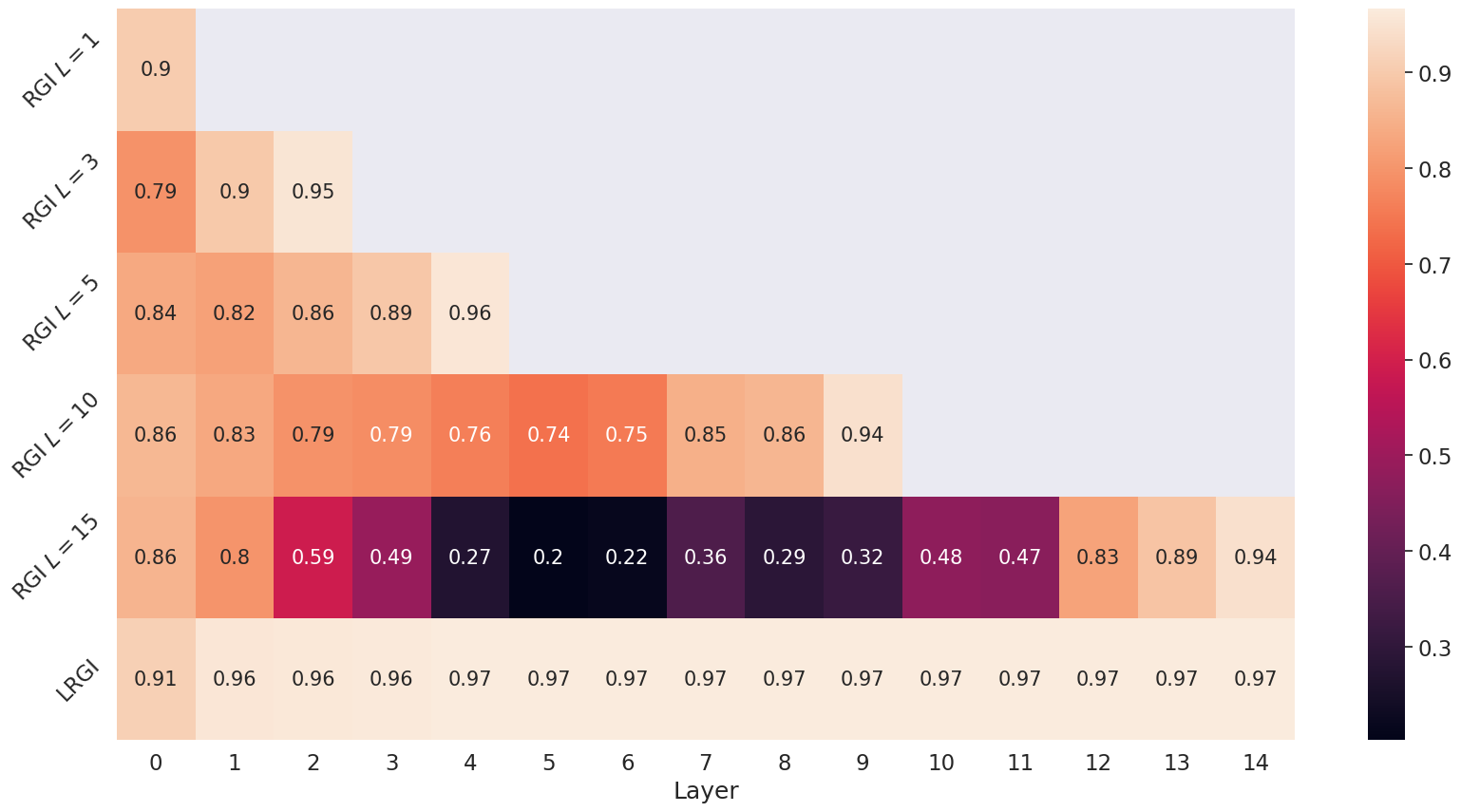}
    \caption{MAD values on the hidden layers of the encoder.}
    \label{fig:depth_oversmoothing}
  \end{subfigure}
  \caption{Effects of depth and width on RGI and LRGI performance and oversmoothing.}
  \small (a) LRGI is able to train deeper and wider models without running into out-of-memory issues due to its memory efficiency. Additionally, it does not suffer from oversmoothing of the node representations and performance is kept, and even improved, when more layers are stacked. On the other hand, while achieving state-of-the-art performance when both the dimensionality and depth are optimal, RGI collapses when the model is too deep. Additionally, the widest model cannot be fit with RGI for deeper models due to memory constraints. (b) MAD values when $f_{\Theta,L}$ is trained with RGI and LRGI for distinct values of $L=\{1, 3, 5, 10, 15 \}$. LRGI is able to avoid oversmoothing and keep a constant MAD value whereas RGI on deep models collapses.
  \label{fig:combined}
\end{figure}

Figure \ref{fig:depth_width_ppi} shows that the downstream task performance of LRGI slightly improves with the depth of the encoder, whereas it drops for the end-to-end RGI when $L>5$. The reason is that the LRGI trained model is able to avoid the oversmoothing of the node representations regardless its depth, as the layer-wise objective promotes the node features to be uncorrelated at each single layer. Figure \ref{fig:depth_oversmoothing} shows an analysis of the oversmoothing of the representations when the GNN encoder is trained with RGI and LRGI, $D=2048$. To quantify the oversmoothing, we employ the mean average (cosine) distance (MAD) measure between the representations of neighbor nodes \cite{chen2020measuring}. When the model is too deep and trained end-to-end, the node representations at the hidden layers are oversmoothed (low MAD values). Note that at the output layer the representations show high MAD values due to the variance-covariance regularization, but as they have collapsed in the inner layers, they are not meaningful and the performance is dropped. Instead, with layer-wise training the MAD values at the output of every layer are constant, so that the oversmoothing effect is avoided and the representations are useful for downstream tasks.

\section{Limitations and future work}
\label{sec:limitations}

In this manuscript, we have presented an algorithm for layer-wise training of graph neural networks in a self-supervised manner. We have shown the computational benefits of layer-wise training with respect to its end-to-end counterpart while keeping a similar performance. Here, we address some of the shortcomings of our algorithm. First of all, layer-wise training increases the number of hyperparameters, as every layer is trained as an independent module so distinct loss weight hyperparameters, for example, could be set. For simplicity, in our experiments we have employed the same hyperparameters for every layer, but the reported performance could be improved with an exhaustive hyperparameter search. Our algorithm is also limited to self-supervised scenarios, as it remains unclear how to train layer-wise using the ground truth label signals in supervised scenarios. We will address this line of research in future work.
\section{Conclusions}
\label{sec:conclusions}
In this work we have presented \textit{Layer-wise Regularized Graph Infomax (LRGI)}, an algorithm to train graph neural networks (GNN) layer by layer in order to overcome the computational challenges of training GNNs end-to-end on large graphs. LRGI's motivation is based on neuroscience and is also self-supervised, avoiding the need of exhaustive annotations. We have shown that our method achieves comparable performance with other state-of-the-art methods for self-supervised learning on graphs, while being much more efficient to train as the space and time complexity are substantially reduced. Therefore, our algorithm enables more sophisticated GNN architectures to be trained without computational constraints.

\bibliographystyle{unsrtnat}
\bibliography{reference}

\begin{thebibliography}{24}
\providecommand{\natexlab}[1]{#1}
\providecommand{\url}[1]{\texttt{#1}}
\expandafter\ifx\csname urlstyle\endcsname\relax
  \providecommand{\doi}[1]{doi: #1}\else
  \providecommand{\doi}{doi: \begingroup \urlstyle{rm}\Url}\fi

\bibitem[Hamilton et~al.(2017)Hamilton, Ying, and Leskovec]{NIPS2017_5dd9db5e}
Will Hamilton, Zhitao Ying, and Jure Leskovec.
\newblock Inductive representation learning on large graphs.
\newblock In I.~Guyon, U.~Von Luxburg, S.~Bengio, H.~Wallach, R.~Fergus,
  S.~Vishwanathan, and R.~Garnett, editors, \emph{Advances in Neural
  Information Processing Systems}, volume~30. Curran Associates, Inc., 2017.
\newblock URL
  \url{https://proceedings.neurips.cc/paper/2017/file/5dd9db5e033da9c6fb5ba83c7a7ebea9-Paper.pdf}.

\bibitem[Pina and Vilaplana(2023)]{pina2023rgi}
Oscar Pina and Verónica Vilaplana.
\newblock Rgi : Regularized graph infomax for self-supervised learning on
  graphs, 2023.

\bibitem[Oord et~al.(2018)Oord, Li, and Vinyals]{oord2018representation}
Aaron van~den Oord, Yazhe Li, and Oriol Vinyals.
\newblock Representation learning with contrastive predictive coding.
\newblock \emph{arXiv preprint arXiv:1807.03748}, 2018.

\bibitem[Hjelm et~al.(2019)Hjelm, Fedorov, Lavoie-Marchildon, Grewal, Bachman,
  Trischler, and Bengio]{hjelm2018learning}
R~Devon Hjelm, Alex Fedorov, Samuel Lavoie-Marchildon, Karan Grewal, Phil
  Bachman, Adam Trischler, and Yoshua Bengio.
\newblock Learning deep representations by mutual information estimation and
  maximization.
\newblock In \emph{International Conference on Learning Representations}, 2019.
\newblock URL \url{https://openreview.net/forum?id=Bklr3j0cKX}.

\bibitem[L{\"o}we et~al.(2019)L{\"o}we, O'Connor, and Veeling]{lowe2019putting}
Sindy L{\"o}we, Peter O'Connor, and Bastiaan Veeling.
\newblock Putting an end to end-to-end: Gradient-isolated learning of
  representations.
\newblock In \emph{Advances in Neural Information Processing Systems}, pages
  3039--3051, 2019.

\bibitem[Veli{\v{c}}kovi{\'{c}} et~al.(2019)Veli{\v{c}}kovi{\'{c}}, Fedus,
  Hamilton, Li{\`{o}}, Bengio, and Hjelm]{velickovic2018deep}
Petar Veli{\v{c}}kovi{\'{c}}, William Fedus, William~L. Hamilton, Pietro
  Li{\`{o}}, Yoshua Bengio, and R~Devon Hjelm.
\newblock {Deep Graph Infomax}.
\newblock In \emph{International Conference on Learning Representations}, 2019.
\newblock URL \url{https://openreview.net/forum?id=rklz9iAcKQ}.

\bibitem[Hassani and Khasahmadi(2020)]{icml2020_1971}
Kaveh Hassani and Amir~Hosein Khasahmadi.
\newblock Contrastive multi-view representation learning on graphs.
\newblock In \emph{Proceedings of International Conference on Machine
  Learning}, pages 3451--3461. 2020.

\bibitem[Peng et~al.(2020)Peng, Huang, Luo, Zheng, Rong, Xu, and
  Huang]{peng2020graph}
Zhen Peng, Wenbing Huang, Minnan Luo, Qinghua Zheng, Yu~Rong, Tingyang Xu, and
  Junzhou Huang.
\newblock {Graph Representation Learning via Graphical Mutual Information
  Maximization}.
\newblock In \emph{Proceedings of The Web Conference}, 2020.
\newblock \doi{https://doi.org/10.1145/3366423.3380112}.

\bibitem[Chen et~al.(2020{\natexlab{a}})Chen, Kornblith, Norouzi, and
  Hinton]{chen2020simple}
Ting Chen, Simon Kornblith, Mohammad Norouzi, and Geoffrey Hinton.
\newblock A simple framework for contrastive learning of visual
  representations.
\newblock \emph{arXiv preprint arXiv:2002.05709}, 2020{\natexlab{a}}.

\bibitem[Zhu et~al.(2020)Zhu, Xu, Yu, Liu, Wu, and Wang]{Zhu:2020vf}
Yanqiao Zhu, Yichen Xu, Feng Yu, Qiang Liu, Shu Wu, and Liang Wang.
\newblock {Deep Graph Contrastive Representation Learning}.
\newblock In \emph{ICML Workshop on Graph Representation Learning and Beyond},
  2020.
\newblock URL \url{http://arxiv.org/abs/2006.04131}.

\bibitem[Thakoor et~al.(2021)Thakoor, Tallec, Azar, Azabou, Dyer, Munos,
  Veličković, and Valko]{thakoor2021bootstrapped}
Shantanu Thakoor, Corentin Tallec, Mohammad~Gheshlaghi Azar, Mehdi Azabou,
  Eva~L. Dyer, Rémi Munos, Petar Veličković, and Michal Valko.
\newblock Large-scale representation learning on graphs via bootstrapping,
  2021.

\bibitem[Bielak et~al.(2021)Bielak, Kajdanowicz, and Chawla]{bielak2021graph}
Piotr Bielak, Tomasz Kajdanowicz, and Nitesh~V. Chawla.
\newblock Graph barlow twins: A self-supervised representation learning
  framework for graphs, 2021.

\bibitem[Zhang et~al.(2021)Zhang, Wu, Yan, Wipf, and
  Philip]{zhang2021canonical}
Hengrui Zhang, Qitian Wu, Junchi Yan, David Wipf, and S~Yu Philip.
\newblock From canonical correlation analysis to self-supervised graph neural
  networks.
\newblock In \emph{Thirty-Fifth Conference on Neural Information Processing
  Systems}, 2021.

\bibitem[Xu et~al.(2021)Xu, Cheng, Luo, Chen, and Zhang]{NEURIPS2021_ff1e68e7}
Dongkuan Xu, Wei Cheng, Dongsheng Luo, Haifeng Chen, and Xiang Zhang.
\newblock Infogcl: Information-aware graph contrastive learning.
\newblock In M.~Ranzato, A.~Beygelzimer, Y.~Dauphin, P.S. Liang, and J.~Wortman
  Vaughan, editors, \emph{Advances in Neural Information Processing Systems},
  volume~34, pages 30414--30425. Curran Associates, Inc., 2021.
\newblock URL
  \url{https://proceedings.neurips.cc/paper/2021/file/ff1e68e74c6b16a1a7b5d958b95e120c-Paper.pdf}.

\bibitem[Bardes et~al.(2022)Bardes, Ponce, and LeCun]{bardes2022vicreg}
Adrien Bardes, Jean Ponce, and Yann LeCun.
\newblock Vicreg: Variance-invariance-covariance regularization for
  self-supervised learning.
\newblock In \emph{ICLR}, 2022.

\bibitem[Hou et~al.(2022)Hou, Liu, Dong, Wang, Tang, et~al.]{hou2022graphmae}
Zhenyu Hou, Xiao Liu, Yuxiao Dong, Chunjie Wang, Jie Tang, et~al.
\newblock Graphmae: Self-supervised masked graph autoencoders.
\newblock \emph{arXiv preprint arXiv:2205.10803}, 2022.

\bibitem[Veličković et~al.(2018)Veličković, Cucurull, Casanova, Romero,
  Liò, and Bengio]{velickovic2018graph}
Petar Veličković, Guillem Cucurull, Arantxa Casanova, Adriana Romero, Pietro
  Liò, and Yoshua Bengio.
\newblock Graph attention networks.
\newblock In \emph{International Conference on Learning Representations}, 2018.
\newblock URL \url{https://openreview.net/forum?id=rJXMpikCZ}.

\bibitem[Grill et~al.(2020)Grill, Strub, Altché, Tallec, Richemond,
  Buchatskaya, Doersch, Pires, Guo, Azar, Piot, Kavukcuoglu, Munos, and
  Valko]{grill2020bootstrap}
Jean-Bastien Grill, Florian Strub, Florent Altché, Corentin Tallec, Pierre~H.
  Richemond, Elena Buchatskaya, Carl Doersch, Bernardo~Avila Pires,
  Zhaohan~Daniel Guo, Mohammad~Gheshlaghi Azar, Bilal Piot, Koray Kavukcuoglu,
  Rémi Munos, and Michal Valko.
\newblock Bootstrap your own latent: A new approach to self-supervised
  learning, 2020.

\bibitem[Chen et~al.(2020{\natexlab{b}})Chen, Lin, Li, Li, Zhou, and
  Sun]{chen2020measuring}
Deli Chen, Yankai Lin, Wei Li, Peng Li, Jie Zhou, and Xu~Sun.
\newblock Measuring and relieving the over-smoothing problem for graph neural
  networks from the topological view.
\newblock In \emph{Proceedings of the AAAI conference on artificial
  intelligence}, volume~34, pages 3438--3445, 2020{\natexlab{b}}.

\bibitem[Paszke et~al.(2017)Paszke, Gross, Chintala, Chanan, Yang, DeVito, Lin,
  Desmaison, Antiga, and Lerer]{paszke2017automatic}
Adam Paszke, Sam Gross, Soumith Chintala, Gregory Chanan, Edward Yang, Zachary
  DeVito, Zeming Lin, Alban Desmaison, Luca Antiga, and Adam Lerer.
\newblock Automatic differentiation in pytorch.
\newblock 2017.

\bibitem[Fey and Lenssen(2019)]{Fey/Lenssen/2019}
Matthias Fey and Jan~E. Lenssen.
\newblock Fast graph representation learning with {PyTorch Geometric}.
\newblock In \emph{ICLR Workshop on Representation Learning on Graphs and
  Manifolds}, 2019.

\bibitem[Zbontar et~al.(2021)Zbontar, Jing, Misra, LeCun, and
  Deny]{zbontar2021barlow}
Jure Zbontar, Li~Jing, Ishan Misra, Yann LeCun, and St{\'e}phane Deny.
\newblock Barlow twins: Self-supervised learning via redundancy reduction.
\newblock \emph{arXiv preprint arXiv:2103.03230}, 2021.

\bibitem[Shwartz-Ziv et~al.(2022)Shwartz-Ziv, Balestriero, and
  LeCun]{https://doi.org/10.48550/arxiv.2207.10081}
Ravid Shwartz-Ziv, Randall Balestriero, and Yann LeCun.
\newblock What do we maximize in self-supervised learning?, 2022.
\newblock URL \url{https://arxiv.org/abs/2207.10081}.

\bibitem[Hua et~al.(2021)Hua, Wang, Xue, Wang, Ren, and Zhao]{hua2021feature}
Tianyu Hua, Wenxiao Wang, Zihui Xue, Yue Wang, Sucheng Ren, and Hang Zhao.
\newblock On feature decorrelation in self-supervised learning.
\newblock \emph{arXiv e-prints}, pages arXiv--2105, 2021.

\end{thebibliography}

\appendix

\section{Implementation details}
All our experiments have been developed and run with PyTorch \cite{paszke2017automatic} and PyTorch Geometric \cite{Fey/Lenssen/2019}. Table \ref{tab:all_hyperparams} shows all the hyperparameters of the experiments. For simplicity and better generalization, have set the architectural ($L$, $D$, $K$) and mini-batching hyperparameters based on existing literature and the training hyperparameters \textit{epochs}, \textit{learning rate} and \textit{weight decay} have been fixed. Note that the number of epochs in PPI is larger as the number of batches is smaller, so that not enough gradient updates would be executed by fixing it to 100. Finally, the loss function hyperparameters ($\lambda_1$,$\lambda_2$,$\lambda_3$) have been set with a small grid search.

\begin{table}[ht]
    \centering
    \caption{Hyperparameters}
    \label{tab:all_hyperparams}
    \begin{adjustbox}{width=1\textwidth}
        \begin{tabular}{l|ccc|ccc|ccc|ccc}
        \toprule
        \textbf{Dataset} & $L$ & $D$ & $K$ & $B$ & $S_L$ & $S_K$ & epochs & learning rate & weight decay & $\lambda_1$ & $\lambda_2$ & $\lambda_3 $\\
        \midrule
        \textbf{PPI} & 3 & 1024 & 1 & 1 (graph) & - & - & 1000 & 1e-4 & 1e-5 & 25 & 25 & 20\\
        \textbf{Reddit} & 2 & 512 & 1 & 128 & 10 & 5 & 100 & 1e-4 & 1e-5 &  50 & 25 & 10\\
        \textbf{ogbn-products} & 3 & 128 & 1 & 512 & 10 & 5 & 100 & 1e-4 & 1e-5 & 25 & 10 & 1 \\
        \bottomrule
        \end{tabular}
    \end{adjustbox}
\vskip 0.1in
\end{table}

\section{Regularized graph infomax (RGI)}
\label{sec:app_rgi}

The mutual information (MI) between two random variables $I(U;V)$ quantifies the amount of information shared between them. It can be lower bounded by the reconstruction error:
\begin{equation}
    \label{eq:mi_bound}
    I(U;V) = H(U) - H(U|V) \geq H(U) - \mathcal{R}(U|V)
\end{equation}
Regularized graph infomax (RGI) \cite{pina2023rgi} trains a graph neural network (GNN) encoder in a self-supervised manner by maximizing the mutual information between local and global node views, $U, V$, respectively. Afterwards, the local views are employed to fit downstream tasks more efficiently. Alternatively, the pre-trained encoder $f_{L,\Theta}$ can be fine-tuned with the task.

Based on Equation \ref{eq:mi_bound}, the maximization of $I(U;V)$ can be addressed with a generative model by minimizing the reconstruction error of $U$ given $V$, $\mathcal{R}(U|V)$. Nonetheless, the entropy of the local views $H(U)$ is not fixed, and directly minimizing $\mathcal{R}(U|V)$ would also decrease the value of $H(U)$, leading to a representation collapse. Consequently, RGI addresses both the maximization of $H(U)$ and the minimization of $\mathcal{R}(U|V)$ in order to maximize the MI between the two views.

\paragraph{Reconstruction error} To minimize $\mathcal{R}(U|V)$, RGI includes a fully-connected network $h_{\phi}$ parameterized by $\phi$, that, for every node $i \in \mathcal{G}$, reconstructs its local view $\mathbf{u}_i$ given the global $\mathbf{v}_i$. The reconstruction is quantified with the square loss:

\begin{equation}
    \label{eq:loss_rec}
    \mathcal{L}_{rec}(U|V) = \frac{1}{N} \sum_{i=1}^{N} \parallel  \mathbf{u}_i - h_{\phi}( \mathbf{v}_i ) \parallel _2^2
\end{equation}

The reconstruction network $h_{\phi}$ is jointly optimized with the GNN encoder $f_{L,\Theta}$ during training, but $h_{\phi}$ is ignored for inference.

\paragraph{Entropy regularization} The maximization of $H(U)$ is tackled by regularizing the covariance matrix of the representation space. The authors follow the variance-covariance regularization proposed in Barlow Twins and VICReg \cite{zbontar2021barlow, bardes2022vicreg, https://doi.org/10.48550/arxiv.2207.10081} to regularize the entropy. Concretely, two main loss terms are defined. Firstly, the variance term maximizes the diagonal elements of $\mathcal{C}$, or variances, to a desired value (i.e. 1):


\begin{equation}
    \mathcal{L}_{var}(U) = \frac{1}{D} \sum_{n=1}^{D} \left ( 1 - \mathcal{C}_{nn} \right )^2
\end{equation}

where $D$ is the number of dimensions of the embedding space. Intuitively, this term avoids the collapse to a constant representation by spreading out the data points. The second term is the covariance minimization, which forces the non-diagonal elements of $\mathcal{C}$ to be close to 0. Consequently, the encoder is encouraged to leverage the whole capacity of the representation space rather than projecting the points to a lower dimensional subspace, also known as dimensional collapse \cite{hua2021feature}:

\begin{equation}
    \mathcal{L}_{cov}(U) = \frac{1}{D} \sum_{n=1}^{D} \sum_{m \neq n} \mathcal{C}_{nm}^2
\end{equation}

For convenience they are kept as two separate loss terms, but their combination is equivalent to minimizing the Frobenius norm of the difference matrix between $\mathcal{C}$ and the identity matrix.

\paragraph{Loss function} The loss function is a weighted combination of the reconstruction error, the variance and the covariance regularization.
 \begin{equation}
     \mathcal{L}_u = \lambda_1 \mathcal{L}_{rec}(U;V) + \lambda_2 \mathcal{L}_{var}(U) + \lambda_3 \mathcal{L}_{cov}(U)
 \end{equation}

where $\lambda_1$, $\lambda_2$, $\lambda_3 \in \mathbb{R}$ are non-learnable weight parameters. In practice, this loss is symmetrized by maximizing $H(V)$ with the entropy regularization applied to the sample covariance of the global views and minimizing $\mathcal{R}(V|U)$, which requires an additional fully-connected network, $h_{\psi}$, that is trained to reconstruct $\mathbf{v}_i$ from $\mathbf{u}_i$:

 \begin{equation}
     \mathcal{L}_v = \lambda_1 \mathcal{L}_{rec}(V;U) + \lambda_2 \mathcal{L}_{var}(V) + \lambda_3 \mathcal{L}_{cov}(V)
 \end{equation}

This symmetrization is carried out because despite the fact that $I(U;V)$ is symmetric, the approximations are not. Finally, the final loss function is a combination of them:

\begin{equation}
    \mathcal{L} = \mathcal{L}_u + \mathcal{L}_v
\end{equation}

\begin{algorithm}[tb]
   \caption{RGI}
   \label{rgi_algo}
\begin{algorithmic}
   \STATE {\bfseries Input:} node features $\mathbf{X}$; adjacency matrix $\mathbf{A}$; backbone $f_{L, \Theta}$; global view function $g_{K}$; reconstruction networks $h_{\phi}$ and $h_{\psi}$.
   \REPEAT
   \STATE // obtain local views
   \STATE $\mathbf{U} = f_{\Theta}({\mathbf{X}}, {\mathbf{A}})$
   \STATE // propagate during K steps
   \STATE $\mathbf{V} = g_{K}({\mathbf{U}, \mathbf{A})}$
   \STATE // reconstruction between views
   \STATE $\mathbf{V}'=h_{\phi}(\mathbf{U})$
   \STATE $\mathbf{U}'=h_{\psi}(\mathbf{V})$
   \STATE // covariance matrices
   \STATE $\mathcal{C}_u= \frac{1}{N} \bar{\mathbf{U}}^T \bar{\mathbf{U}} $
   \STATE $\mathcal{C}_v     = \frac{1}{N} \bar{\mathbf{V}}^T \bar{\mathbf{V}} $
   \STATE // loss function
   \STATE $\mathcal{L}_{rec} =  \parallel \mathbf{U} - \mathbf{U}' \parallel^2_F +  \parallel \mathbf{V} - \mathbf{V}' \parallel^2_F $
   \STATE $\mathcal{L}_{var} = (1-$diag$(\mathcal{C}_u))^2 + (1-$diag$(\mathcal{C}_v))^2$
   \STATE $\mathcal{L}_{cov} = ($off-diag$(\mathcal{C}_u))^2 + ($off-diag$(\mathcal{C}_v))^2$
   \STATE $\mathcal{L} = \lambda_1 \mathcal{L}_{rec} + \lambda_2 \mathcal{L}_{var} + \lambda_3 \mathcal{L}_{cov} $
   \STATE // update parameters
   \STATE $\Theta, \phi, \psi \leftarrow \bigtriangledown_{\Theta, \phi, \psi} \mathcal{L}$
   \UNTIL{convergence}
   \STATE {\bfseries return} $\mathbf{U}$
\end{algorithmic}
\end{algorithm}

\end{document}